\documentclass[a4paper]{article}

\usepackage{INTERSPEECH2015}

\usepackage{graphicx}
\usepackage{amssymb,amsmath,bm}
\usepackage{textcomp}
\usepackage{multirow}
\usepackage{dsfont}

\ninept

\title{Learning from Real Users: Rating Dialogue Success with Neural Networks for Reinforcement Learning in Spoken Dialogue Systems}

\makeatletter
\def\name#1{\gdef\@name{#1\\}}
\makeatother \name{{\em Pei-Hao Su, David Vandyke, Milica Ga{\v s}i{\' c}, Dongho Kim,} \\
{\em Nikola Mrk{\v s}i{\' c}, Tsung-Hsien Wen and Steve Young}}

\address{Department of Engineering, University of Cambridge, Cambridge, UK \\
  {\small \tt \{phs26, djv27, mg436, dk449, nm480, thw28, sjy\}@cam.ac.uk}
}

\begin{document}
\maketitle

\begin{abstract}

\noindent To train a statistical spoken dialogue system (SDS) it is essential that an accurate method for measuring task success is available. To date training has relied on presenting a task to either simulated or paid users and inferring the dialogue's success by observing whether this presented task was achieved or not. Our aim however is to be able to learn from real users acting under their own volition, in which case it is non-trivial to rate the success as any prior knowledge of the task is simply unavailable. User feedback may be utilised but has been found to be inconsistent. Hence, here we present two neural network models that evaluate a sequence of turn-level features to rate the success of a dialogue. Importantly these models make no use of any prior knowledge of the user's task. The models are trained on dialogues generated by a simulated user and the best model is then used to train a policy on-line which is shown to perform at least as well as a baseline system using prior knowledge of the user's task. We note that the models should also be of interest for evaluating SDS and for monitoring a dialogue in rule-based SDS. 

\end{abstract}


\section{Introduction}
\label{sec:intro}

The dialogue manager is the core component of a spoken dialogue system (SDS). It controls the interaction between the system and the user, and is central to the overall quality of the user experience. Casting an SDS as a partially observable Markov decision process (POMDP) has been shown to be beneficial by allowing the dialogue manager to be optimised to plan and act under the uncertainty created by noisy speech recognition and semantic decoding \cite{POMDP_williams, POMDP-review}. The POMDP policy dictating the actions taken by the SDS is trained in an episodic reinforcement learning (RL) framework \cite{RL} whereby the agent receives a reinforcement signal after each dialogue (episode) reflecting how well it performed.


The goal of this paper is to demonstrate that an SDS can be trained via interactions with real users where no direct knowledge of the user's goals is known at any point in the dialogue. In all previous works the training of an SDS has been done with either recruited subjects \cite{GPRL, GasicKTBHSTY14} who are presented with a pre-defined task to complete, or via simulated users \cite{Lemon07machinelearning, daubigney2012comprehensive, levin_POMDP, userSim, userSim2} who randomly sample a goal over the specific ontology. In both cases, the specific prior knowledge of the user's goal is used to calculate an objective measure (\textit{Obj}) of whether the SDS completed the task or not. In real world systems prior knowledge of the user's goal is simply not available, making any calculation of an `objective' measure nearly impossible\footnotemark{}. Knowledge of task success or failure is essential however for training an SDS.
\newpage

\footnotetext{We note that this is not a problem faced in training agents in many common POMDP tasks: episode success in grid-worlds, games or pole-balancing is well defined and easily computed \cite{RL}. In comparison, dialogue is an ill-posed problem for which it is non-trivial to classify the success of an episode when there is no prior knowledge of the user's goal. There is even ambiguity as to what the label success means for a dialogue. Our definition of success is based on the performance of the dialogue agent, specifically whether it provided all of the information asked of it for a domain entity satisfying the users constraints, \textit{e.g.} the \textit{phone number} for a \textit{cheap} restaurant in the \textit{north}.}

One approach to this problem is to ask the user for feedback at the completion of each dialogue. Yang {\it et al}. \cite{yang2012predicting} proposed using collaborative filtering to infer user preferences given a set of user-rated dialogues. However these ratings were very noisy \cite{Daniel_RSS_2014} which lead to slow learning and poor policies \cite{milica_real_users}. Also in real-world  systems it is not clear that a user would be cooperative enough to provide feedback once the dialogue is completed.

Other research related to this problem includes the PARADISE framework \cite{walker1997paradise} presented by Walker \textit{et al.} for evaluating a dialogue, where a linear function of task completion and predefined dialogue costs were used for inferring user satisfaction. However, as noted above, task completion is not directly computable with real users and concerns relating to the theoretical motivation of the model have also be raised \cite{larsen_2003}. A framework that does potentially enable the training of SDS with real users was presented by Asri \textit{et al.} \cite{asri_rfl, TCTL}, whereby a reward function was learnt over a summary state space based on dialogue data labelled by experts for task success.  However, no attempt was made to learn a policy with real users.

When training an SDS with paid users given specific tasks, a common issue is that they are not motivated by a real information need.
As a consequence, they often\footnotemark{} fail to follow exactly the presented goal, resulting in \textit{Obj}=failure even though the SDS may have actually provided everything asked of it. In order not to penalise the SDS by learning with such dialogues we have previously also asked the user for their opinion of whether they achieved the task goals thereby obtaining a subjective success rating (\textit{Subj}).  Then for policy learning,  only those dialogues for which \textit{Obj}=\textit{Subj} \cite{milica_icassp13} are used, the remainder being discarded.   
With real users it is not possible however to calculate \textit{Obj} since the true goal of the user can not be known. It is therefore essential to find effective methods for computing rewards with real users when the underlying task is unknown.
\footnotetext{This case occurs in our experience at least 20\% of the time \cite{milica_real_users}.}



This paper investigates the use of neural networks to rate task success automatically on-line by tracking the dialogue as it evolves.
In Section \ref{sec:models}, two types of neural networks are described, recurrent neural nets (RNNs) and convolutional neural nets (CNNs), 
and the choice of features used to track the dialogue are discussed along with the different types of predictions the models are trained to produce.  
The experimental evaluation is then presented in Section \ref{sec:exp}.  Two performance metrics are computed to evaluate the trained NN models: accuracy in estimated task success and the root mean square error in estimating the reward function.   Performance in on-line learning with (paid) users is then assessed and the effectiveness of the neural network-based reward rating is demonstrated.
Finally, conclusions are presented in Section \ref{sec:conclusions}.


\section{Neural network dialogue classification}
\label{sec:models}
Two types of neural network (NN) models were investigated for determining the final reward given to the reinforcement learning agent. The structures of these models are described in sub-sections \ref{sub:rnn}, \ref{sub:cnn} and \ref{sub:output}. First though we discuss their shared feature inputs and training data.

\subsection{Training data, dialogue features and generalisation}
\label{sub:feat}

The data used to train all models was collected by training several Gaussian Process policies \cite{GPRL} from scratch with an agenda-based simulated user \cite{userSim, userSim2}. The labels of success or failure for each dialogue were computed based on an objective criteria of whether or not the agent met the simulated users' goals generated at the start of each dialogue. The reinforcement signals used during policy training were simply to give a -1 reward at each turn to promote speed, and a final reward of +20 at completion if the dialogue was successful, otherwise 0. The return (cumulative reward) $R$ is therefore calculated as:
\begin{equation}
R = 20\times\mathds{1}_{success}- N
\label{eqn:return}
\end{equation}
where $N$ is the number of turns in a dialogue and $\mathds{1}_{success}$ is an indicator function for success.

\begin{figure}[h]
\centerline{\includegraphics[scale=0.45]{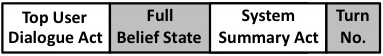}}
\vspace{-2mm}
\caption{{Feature vector ${\bf f}_t$ extracted at each turn $t$.}} 
\vspace{-1mm}
\label{fig:feature}
\end{figure}

For all models, a domain specific feature vector was extracted at each turn\footnotemark{} consisting of the following concatenated sections: one-hot encoding of the user's top-ranked dialogue act, the real-valued belief state vector formed by concatenating the distributions over all goal, method and history variables \cite{BUDS}, one-hot encoding of the summary system action, and  the turn number. This is shown in Figure \ref{fig:feature}. 
\footnotetext{Turn here means system + user exchange.}

\vspace{-3mm}

\noindent
This form of feature vector was motivated by considering the primary information a human would require to read a transcription and rate the success of the dialogue. The inclusion of the full belief vector, plus user and systems actions makes this feature vector domain and system dependent.

The goal with these NN models is to enable policy learning with real users by not requiring any prior knowledge of the users' goal. Their rating predictions are used directly to provide the RL feedback to the dialogue agent. Hence they should consider the information requested by the user over the whole dialogue and ideally evaluate whether the policy provided everything that was asked for or not. It is expected that by training the NN models on data from the simulated users evaluated by the objective measure, they will generalise to be able to provide this ideal rating when assessing dialogues with real users whose goals are not known (and hence the objective assessment can not be calculated). The reason to expect the models to generalise in this way is that the simulated users have predefined tasks and inform the system meticulously about all of them.
Hence, the objective measure of task completion indicates exactly whether or not the system provided the information requested of it. Therefore by training on these supervised learning pairs of data generated by the simulated user and ratings provided by the objective measure, the resulting NN predictive model should be a good detector of whether or not the system provided what the user requested from it. This is the desired indicator of the system's behaviour and a good reinforcement signal for policy learning.



Dialogues of course vary in their total number of turns. By extracting this feature vector at each turn a variable size set is obtained for each dialogue. The two NN models we investigate both make a single prediction for the whole dialogue, but do so in different ways, in particular with respect to how they handle this variable length sequence.

\subsection{Recurrent neural network model}
\label{sub:rnn}

The recurrent neural network (RNN) model \cite{Lukoševičius2009127} is a subclass of neural network that has feedback connections from one time step to another. The ability to succinctly capture and retain history information makes it suitable for modelling sequential data with temporal dependencies. It has been shown to be successful in various natural language processing tasks such as language modelling  \cite{mikolov2010recurrent, mikolov2011extensions, KarpathyF14} and spoken language understanding (SLU) \cite{RNNonSLU}.

\begin{figure}[t]
\centerline{\includegraphics[width=75mm]{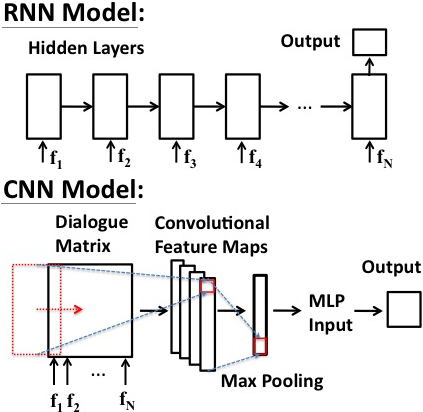}}
\caption{{Schematic of the two NN predictive models. An unrolled view of the RNN (\textit{top}) and CNN model (\textit{bottom}). The feature vectors extracted at turns $t=1,\dots,N$ are labelled ${\bf f}_t$.}} 
\vspace{-4mm}
\label{fig:models}
\end{figure}

Here the RNN model is adopted to manage the variable length of each dialogue by simply updating its hidden layer ${\bf h}_t$ with the input feature vectors ${\bf f}_t$ at each turn $t$. Once the dialogue ends the hidden layer is then connected to an output layer to make a single prediction of the whole dialogue as depicted in the top half of Figure \ref{fig:models}. 


\subsection{Convolutional neural network model}
\label{sub:cnn}
Also investigated was a convolutional neural network (CNN) \cite{CNN} which has been successfully used  for image classification \cite{Lecun_cnn} and on sequential modelling problems such as sentiment labelling of sentences \cite{CNN_sent}. Here the CNN makes predictions by considering the whole dialogue as a matrix formed by appending turn based feature vectors. On completion of the dialogue, a convolutional filter of size $(F,W)$, where $F$ is the turn based feature dimension and $W$ is a width across time, is applied in a narrow convolution across the dialogue matrix. Multiple filters are used, each of which creates its own feature map. A max-pooling operation then reduces each of the feature map vectors to a scalar. Finally, the resulting scalars are concatenated and feed into a standard multi-layer perceptron (MLP), which may consist of multiple layers. This process is shown in the bottom half of Figure \ref{fig:models}, where 4 feature maps are employed.


For the CNN, the mapping of the variable size input to a fixed size is provided by the pooling operation applied to the feature map outputs. The dialogue matrix is  padded with $W-1$ zero vectors on each side to allow a narrow convolution to always be performed (even if the dialogue has only 1 turn). Importantly this also allows the convolutional filter to move across time (turns) and consider turn sequences of differing lengths.


\subsection{RNN \& CNN shared output layer}
\label{sub:output}

The RNN and CNN models share the same network structure in their final layer, and this structure is determined by the choice of supervised training targets, of which three types were considered, all derived from the described data. 


\textit{1) In the first case} the NN models are classifiers which are trained to predict the \textit{Obj} success or failure label for each dialogue. The targets are $\{0,1\}$ and the final layer of the  NN models outputs a scalar through a sigmoid activation function and is trained with a cross-entropy loss. The outputs from this network is a probability $p$ that the dialogue is a success, and the hard class label predicted by the model is taken as 1 if $p>0.5$, else 0. This hard label is used to determine whether to give a final reward of +20 during policy learning, as per Eqn. (\ref{eqn:return}). 

In the other two cases, given that our goal is to provide the final RL reward for policy learning, we also investigate predicting this reward directly.

\textit{2) The second case} is a multiclass classification problem where the class labels are integers representing the possible returns for the whole dialogue. The number of different returns possible with Eqn. (\ref{eqn:return}) is constrained by setting a maximum number of allowable turns for a dialogue. A softmax activation is used in the final layer of the NN models with a cross-entropy loss. 
The one-hot encoding of the target distributions are convolved by a discrete Gaussian kernel in order to smooth and reduce the magnitude of the return prediction errors. 


\textit{3) The third case} is a regression problem with the actual return value used as the training target. The final layer of the NN models have no non-linearity (activation) and the whole model is trained with a mean-square-error (MSE) loss function. During policy learning with cases 2 \& 3 a per-turn penalty of 0 would be used, since these models predict the return rather than the final reward and so implicitly include the total number of  turns penalty in the predicted return.


\section{Experiments}
\label{sec:exp}
\subsection{Domain and shared SDS components}
\label{sub:domain}


In all experiments the Cambridge restaurant domain was used, which consists of approximately 150 venues each having 6 attributes (slots) of which 3 can be used by the system to constrain the search and the remaining 3 are informable properties once a database entity has been found. 

The shared core components of the SDS used over all experiments were a domain independent ASR, a confusion network (CNet) semantic input decoder \cite{CNET}, the BUDS \cite{BUDS} belief state tracker that factorises the dialogue state using a dynamic Bayesian network and a template based NLG of the systems semantic actions. All policies are trained by GP-SARSA \cite{GPRL} and the summary action space contains 20 actions. 


With this ontology, the number of elements in each of the four segments of the feature vector used by the NN models were 21, 575, 20, 1 respectively for the user act, full belief state, system act and turn number. This resulted in a vector of $F=617$ components at each turn. The turn number was expressed as a percentage of the maximum number of allowed turns, here 30. The one-hot user dialogue act encoding was formed by taking only the most likely user act estimated by the CNet decoder.

\subsection{Results: Neural network training}

\begin{figure*}[t]
\centerline{\includegraphics[width=150mm]{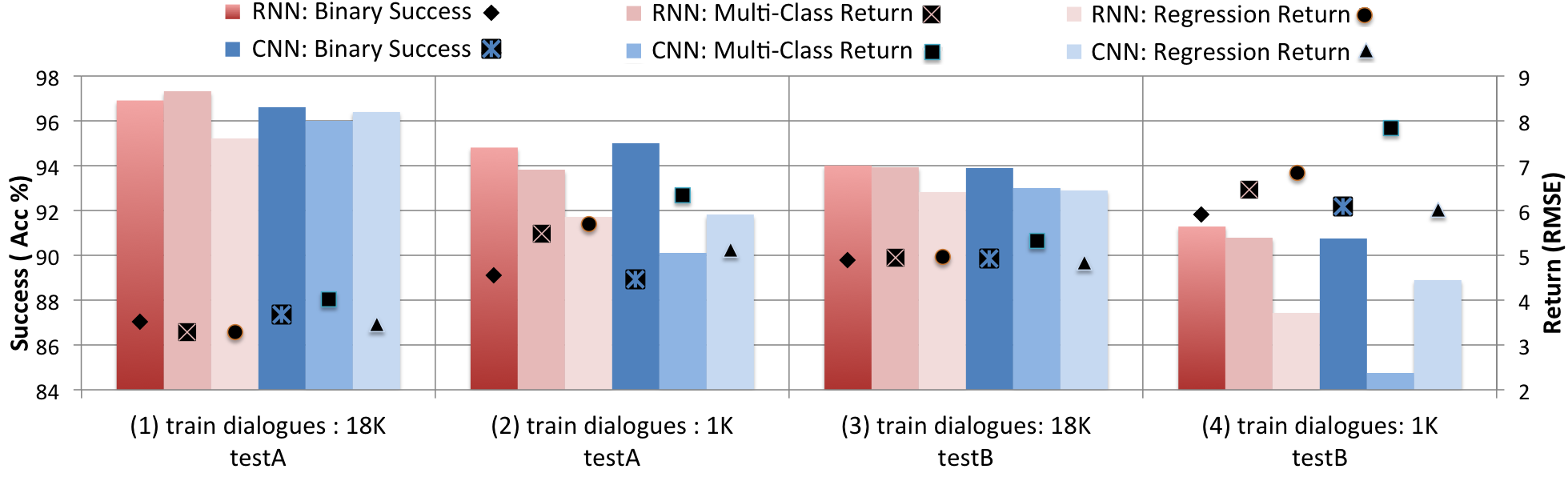}}
\vspace{-4mm}
\caption{{NN model training results: Prediction of RNN and CNN models trained on 18K and 1K dialogues and tested on sets \textit{testA} and \textit{testB} (see text). Results of success/failure label accuracy (left axis) are represented as bars, and RMSE (right axis) as scatters. }} 
\label{fig:NN_training}
\end{figure*}

In this section results of training the two NN models\footnotemark{} on the simulated user \cite{userSim} dialogues scored by the \textit{Obj} measure are presented. Two training sets were used consisting of 18K and 1K dialogues. In all cases a separate validation set  consisting of 1K dialogues was used for controlling overfitting. Training and validation sets were approximately balanced regarding objective success/failure labels and collected at a 15\% semantic error rate (SER). Prediction results are shown in Figure \ref{fig:NN_training} on two test sets; \textit{testA:} 1K dialogues, balanced regarding objective labels, at 15\% SER and \textit{testB:} 12K dialogues,  containing 3 GP policies trained from scratch on 1000 dialogues, collected at an SER of $0,15,30$ and $45$ as the data occurred (\textit{i.e.} with no balancing regarding labels). 

\footnotetext{All NN models were implemented using the Theano library \cite{bergstra+al:2010-scipy, Bastien-Theano-2012}. The RNN hidden layer used 300 units with sigmoid activations for all cases. The CNN created 50 feature maps with filters of width $W=30$, and a 2 layer FFNN where the size of the 1st layer was 300 in \textit{case 2} and 50 otherwise. Stochastic gradient descent (per dialogue) was used for training.}


We used three different targets (cost functions) as described in section \ref{sub:output} to train both the RNN and CNN models. Eqn. (\ref{eqn:return}) was used to calculate the return from the binary success classification (case 1 in \ref{sub:output}); for cases 2 and 3 the success label was inferred from Eqn. (\ref{eqn:return}). The results are depicted in Figure \ref{fig:NN_training}, where the left y-axis is the success classification accuracy (bar plot), and the right y-axis is the root-mean-square-error (RMSE) of the return (scatter plot).

We see that the RNN outperformed the CNN in most cases. When using the large training set (18K, sub-figures 1 \& 3) all models obtained over 93\% success label accuracy while the RNN more accurately estimated the return, getting within $\sim\pm3$ of the objective return targets on \textit{testA} and within $\sim\pm5$ on \textit{testB}. Without a simulated user it may not be possible to access 18K labelled training dialogues so results are also presented when training the models (with exactly the same structures) on only 1K dialogues.  Sub-figures 2 \& 4 show that the models are reasonably robust to this large reduction in the amount of training data, with the binary classification models being the most accurate and again the RNN  outperforming the CNN. 

These results give confidence that the NN models, sequentially evaluating turn level features, are able to serve as good dialogue success detectors. The results on set \textit{testB} also show that the models can perform well in environments with varying error rates as would be encountered in real operating environments.

\subsection{Results: On-line policy training with the RNN model}


Based on the above results, the binary  RNN classification model was selected for training policies on-line. Two systems were trained on-line by users recruited via Amazon Mechanical Turk\footnotemark{}.  Firstly, a baseline system was trained which used knowledge of the set tasks to compute the reward as described in Section~\ref{sec:intro}, and secondly a system was trained using only the RNN to compute the reward signal.  Three policies were trained for each system, then averaged to reduce noise.
Learning began from a random policy in all cases. 


\footnotetext{Although our motivation is to train with real users and the NN models we have introduced now enable this, we are restricted here to using Mechanical Turkers since we do not have an actual service or product to attract real users to.}

\begin{figure}[h]
\centerline{\includegraphics[width=75mm]{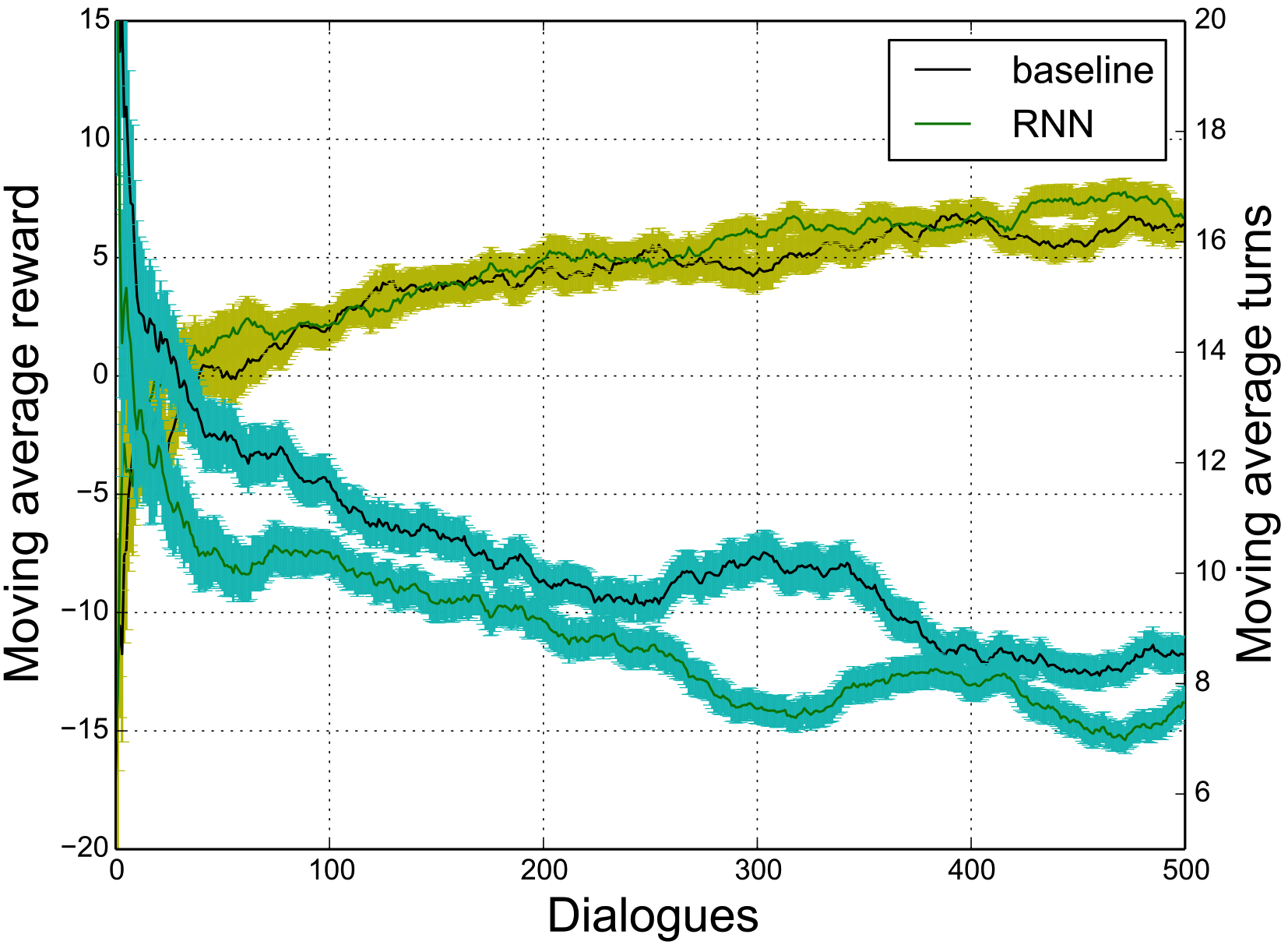}}
\caption{{Learning curve with reward and number of turns during on-line policy optimisation. The baseline system (black line) updates the policy only when the \textit{Subj} and \textit{Obj} measures agreed. The green line shows training under the RNN dialogue success predictor. Yellow and blue lines are standard errors.}} 
\vspace{-3mm}
\label{fig:online_reward}
\end{figure}

Figure \ref{fig:online_reward} shows the on-line learning curve of the reward and number of turns when training the systems with 500 dialogues. For both plots,  the moving average was calculated using a window of 100 dialogues and each result was the average of the three policies in order to reduce noise. It can be seen that the RNN system was able to learn at least as good a policy as the baseline system. Further, the baseline system actually required $\sim850$ dialogues (due to discarding cases where \textit{Obj}$\neq$\textit{Subj}), while the RNN system used every dialogue  and was therefore more efficient and less costly. 


In order to evaluate the resulting policies, we collected a further 600 dialogues, turning off policy learning and asking the Mechanical Turkers to rate, in addition to \textit{Subj}, the quality of the dialogue by answering the question ``\textit{Do you think this dialogue was successful?}" on a 6-point Likert scale. Each of the 3 policies trained for the baseline and RNN systems received 100 dialogues and the average quality rating (interpreted as a number between 0 and 5) is shown in Table \ref{table:eval} along with one standard error. 
We report only the quality and \textit{Subj} since the \textit{Obj} can be misleading due to Turkers not explicitly following the task, as highlighted in Section \ref{sec:intro}. 
The results indicate that the RNN dialogue success classifier was able to train a policy at least as well as the baseline system even though the baseline was trained via direct use of the prior knowledge of the users goal and selected only dialogues where \textit{Obj}=\textit{Subj} to learn from.



\vspace{-5mm}
\begin{table}[h]
\caption{{Subjective evaluations of the trained baseline and RNN policies. Quality: 6-point Likert scale, \textit{Subj}: binary rating. }} 
\begin{center}
\begin{tabular}{l|c|c}
     & baseline & RNN   \\ \hline
Quality (0-5)    & 3.77 $\pm$ 0.087    &  3.94 $\pm$ 0.068  \\ \hline
\textit{Subj} (\%)    & 84.9 $\pm$ 2.2    &  89.5 $\pm$ 1.7  \\ \hline
\end{tabular}
\end{center}
\label{table:eval}
\vspace{-5mm}
\end{table}

\section{Conclusions}
\label{sec:conclusions}


This paper has investigated the use of neural networks for rating success in a spoken dialogue system.   Both RNNs and CNNs were shown to be capable of
good performance when substantial training data is available, but RNNs were more robust when training data was limited.
When compared to a baseline (which used prior knowledge of the users goal) for on-line policy learning with real users, the RNN delivered slightly improved performance suggesting that this
approach does provide a way of training real-world systems on-line with users whose goals are unknown.


Currently work is focused on investigating less domain specific features, the dependence on the simulated user, transferring the RNN models to new domains, and using them for reward shaping \cite{socialRS} to speed up policy learning. We note finally that the models should also be helpful for rule based SDS to adjust behaviour or know when to hand control from the computer agent to a human to retrieve a failing dialogue, and for evaluation of SDS generally. 

\section{Acknowledgement}
\label{sec:acknowledgement}
P.-H. Su is supported by Cambridge Trust and the Ministry of Education, Taiwan. 
D. Vandyke and T.-H. Wen are supported by Toshiba Research Europe Ltd, Cambridge Research Lab.

\newpage
\clearpage
\eightpt
\bibliographystyle{IEEEtran}
\bibliography{ref.bib}

\end{document}